\documentclass[letterpaper, 10 pt, conference]{ieeeconf}  %

\IEEEoverridecommandlockouts                              %

\usepackage{graphics} %
\usepackage{graphicx} %
\usepackage{amsmath} %
\usepackage{booktabs}
\usepackage{hyperref}
\usepackage{fancyhdr}
\usepackage{ifthen}

\newcommand{\version}[1]
{
  \ifthenelse{\equal{#1}{ARXIV}}{\thispagestyle{fancy}}{}
  \ifthenelse{\equal{#1}{WORKSHOP}}{\thispagestyle{plain}}{}
}

\title{\LARGE \bf
Single-View Place Recognition under Seasonal Changes
}

\author{Daniel Olid,  Jos\'e M. F\'acil and Javier Civera%
\thanks{This work was partially supported by the Spanish government (project DPI2015-67275) and the Arag\'on regional government (Grupo DGA-T45\_17R/FSE}%
\thanks{The authors are with the I3A, Universidad de Zaragoza, Spain \newline
        {\tt\small danielolid94@yahoo.es, jmfacil@unizar.es, \newline jcivera@unizar.es}}%
}

\begin{document}

\maketitle

\version{ARXIV} %
\pagestyle{plain}

\begin{abstract}

Single-view place recognition, that we can define as finding an image that corresponds to the same place as a given query image, is a key capability for autonomous navigation and mapping. Although there has been a considerable amount of research in the topic, the high degree of image variability (with viewpoint, illumination or occlusions for example) makes it a research challenge. 

One of the particular challenges, that we address in this work, is weather variation. Seasonal changes can produce drastic appearance changes, that classic low-level features do not model properly. Our contributions in this paper are twofold. First we pre-process and propose a partition for the Nordland dataset, frequently used for place recognition research without consensus on the partitions. And second, we evaluate several neural network architectures such as pre-trained, siamese and triplet for this problem. Our best results outperform the state of the art of the field. A video showing our results can be found in \url{https://youtu.be/VrlxsYZoHDM}. The partitioned version of the Nordland dataset at \mbox{\url{http://webdiis.unizar.es/~jmfacil/pr-nordland/}}.

\end{abstract}

\section{Introduction}
\label{cap:introduccion}
\begin{figure}[t!]
	\centering
	\includegraphics[width=8cm]{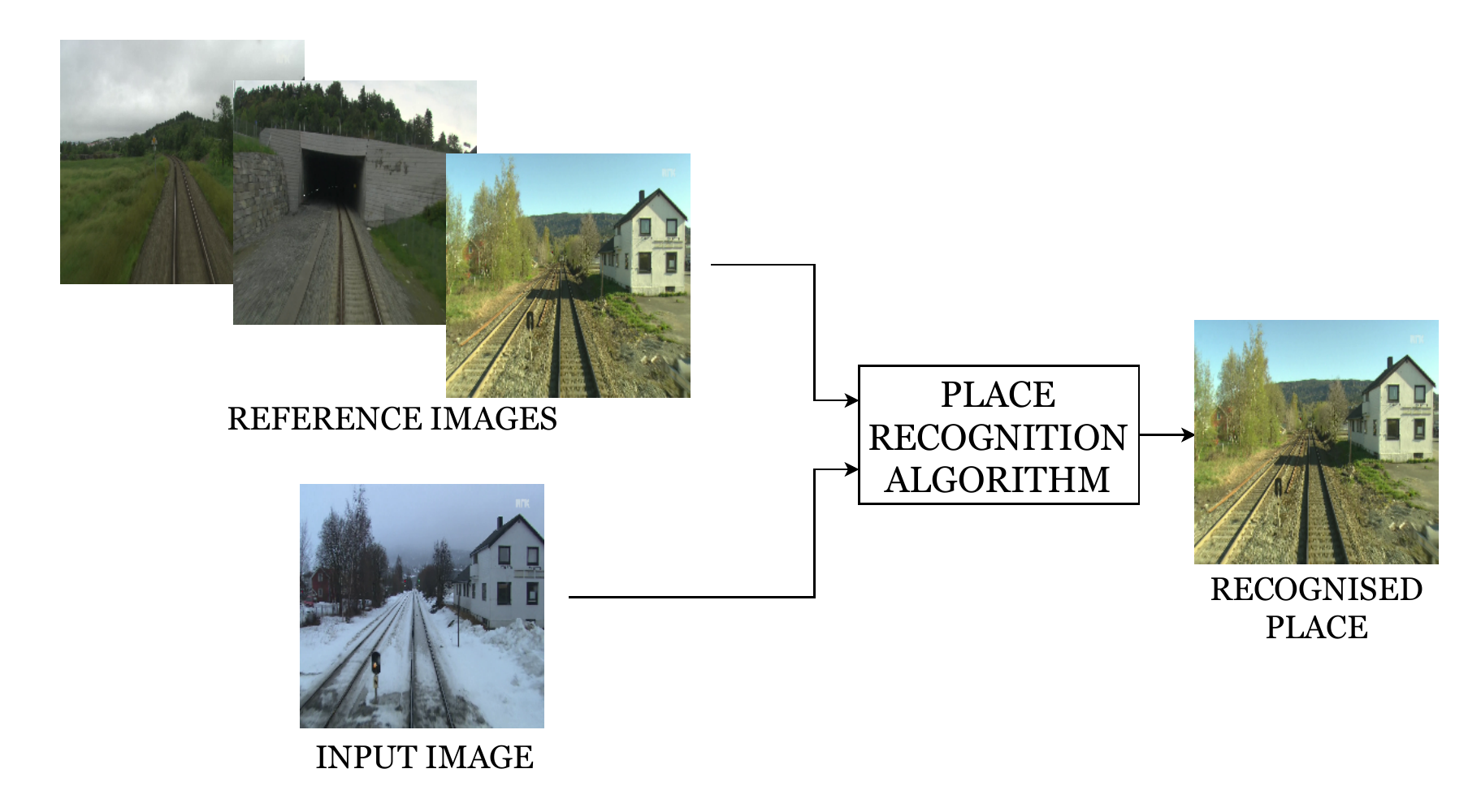}
	\caption{\label{img:bloquesreconocedor}
    Place recognition overview. The inputs are two; a database of images taken in different places, and query view imaging the place to recognize. The output is an image of the database showing the place of the query image.}
\end{figure}

Visual place recognition consists on, having a query image, retrieving from a database another image that corresponds to the same place, see Fig. \ref{img:bloquesreconocedor}. Place recognition plays a relevant role in several applications, \textit{e.g.} mobile robotics. To name a few, place recognition can be used for topological mapping \cite{garcia2017hierarchical}, for loop closure and drift removal in geometric mapping \cite{mur2015orb}, and for learning scene dynamics in lifelong localization and mapping \cite{dymczyk2015gist}. 

Place recognition for robotics presents multiple challenges. For example, most of the times the places databases are huge and the retrieval time is constrained by the real-time operation of robots. Another relevant challenge, which is the one we will address in this paper, is the variability in the visual appearance of the places. The appearance variations might have different sources: viewpoint and illumination changes, occlusions and scene dynamics. 

The appearance changes coming from different viewpoints and illumination conditions, assuming a static scene, have been addressed quite successfully. Local point features (e.g., SIFT, SURF and ORB), based on image gradients, show a high repeatability and descriptor invariance to moderate levels of illumination and viewpoint changes. Checking the geometric and sequential compatibility of such local features can improve even further their robustness \cite{galvez2012bags}. Global image features have been also used for place recognition \cite{sunderhauf2011brief,murillo2013localization}, showing better scalability but lower performance under viewpoint changes or occlusions.

\begin{figure}[t!]
	\centering
	\includegraphics[width=7cm]{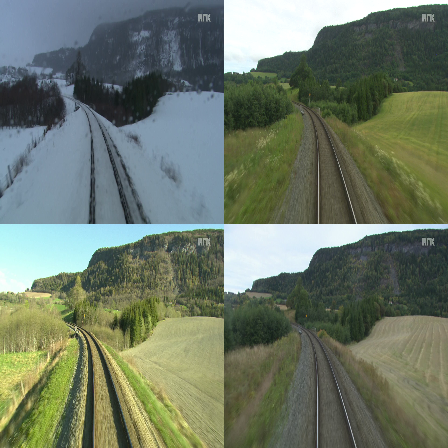}
	\caption{\label{img:nordlandseasons}
    Images from the same place in different seasons. From top-left and clockwise: winter, summer, fall and spring. Notice the appearance change due to different weather conditions. The images have been extracted from the videos of the Nordland dataset.
    }
\end{figure}

The classical approaches based on hand-designed low-level features are, however, limited for the representation of dynamic scene changes. There has been several works aiming at designing descriptors with higher invariance to certain transformations, either based on models (e.g., \cite{simo2015dali}) or based on learning from data (e.g., \cite{verdie2015tilde}). The most recent approaches use Convolutional Neural Networks (CNNs), due to their higher potential to learn image patterns. In this work we explore the use of CNNs for place recognition in the particular case of seasonal changes. Our specific contributions over the state of the art are:

\begin{itemize}
	\item We have trained a weather-invariant place recognition method, based on CNNs. We use CNNs to extract image descriptors, that we compare using the Euclidean distance. Fig. \ref{img:nordlandseasons} depicts some of the weather variations considered.
	\item We have designed a dataset using images extracted from the Nordland videos \cite{NRK}. We propose our Nordland dataset partition as a common framework for evaluating place recognition.
	\item We have compared our results in the Nordland dataset against other state of the art techniques. Our method is capable of correctly recognizing $98\%$ of the input places in $80$km routes under favorable conditions and $86\%$ under drastic appearance changes like the ones occurring between summer and winter.
\end{itemize}

The rest of this paper is structured as follows. Section \ref{sec:trabajo_relacionado} analyzes the related work in place recognition. Section \ref{sec:dataset} explains the development of the dataset. Section \ref{sec:arquitecturas_neuronales} introduces the neural network architectures that we have used. Section \ref{sec:resultados} presents our results. Finally, in section \ref{sec:conclusiones} we summarize our conclusions.

\section{Related work}
\label{sec:trabajo_relacionado}
The most common approaches to place recognition are based on local image features using classic extractors and descriptors. Two of the most relevant among these techniques are FAB-MAP \cite{cummins2008fab} and DBoW \cite{galvez2012bags}. The performance of these algorithms is excellent for moderate viewpoint and illumination changes, but it decreases for other types of appearance variations. 

An alternative approximation consists in using neural networks as feature extractors. S\"underhauf \textit{et al.} analyzed in \cite{DBLP:journals/corr/SunderhaufDSUM15} the use of neural networks for the purpose of place recognition with promising results. \cite{DBLP:journals/corr/SunderhaufDSUM15, neubert2015local} and \cite{gout2017evaluation} were the first ones to use neural networks for this purpose but \cite{gomez2015training} and \cite{chen2017deep} were the first ones to specifically train neural architectures to attack this problem. There is no consensus on what kind of architecture is better for this task.

In this work we compare three different techniques that can be considered state of the art in place recognition: Unsupervised linear learning techniques for visual place recognition \cite{lowry2016supervised}, deep learning features at scale for visual place recognition \cite{chen2017deep} and CNN for appearance-invariant place recognition \cite{gomez2015training}.

The first method \cite{lowry2016supervised} applied principal components analysis to reduce the dimensionality of the feature vector, eliminating the dimensions that are affected by appearance changes. The second method \cite{gomez2015training} used a triplet neural network architecture to fine-tune a pre-trained model and improve the robustness of the extracted features. Their network learned to map images to a vector space where euclidean distance represents similarity. The third method \cite{chen2017deep} trained a deep neural network to classify the place that appeared in a dataset of images taken from surveillance cameras.

This work develops a technique similar to the one implemented in \cite{gomez2015training}. As a novelty, we also train siamese neural networks and consider different pre-trained networks. 

\begin{figure}[t!]
	\centering
	\includegraphics[width=.95\linewidth]{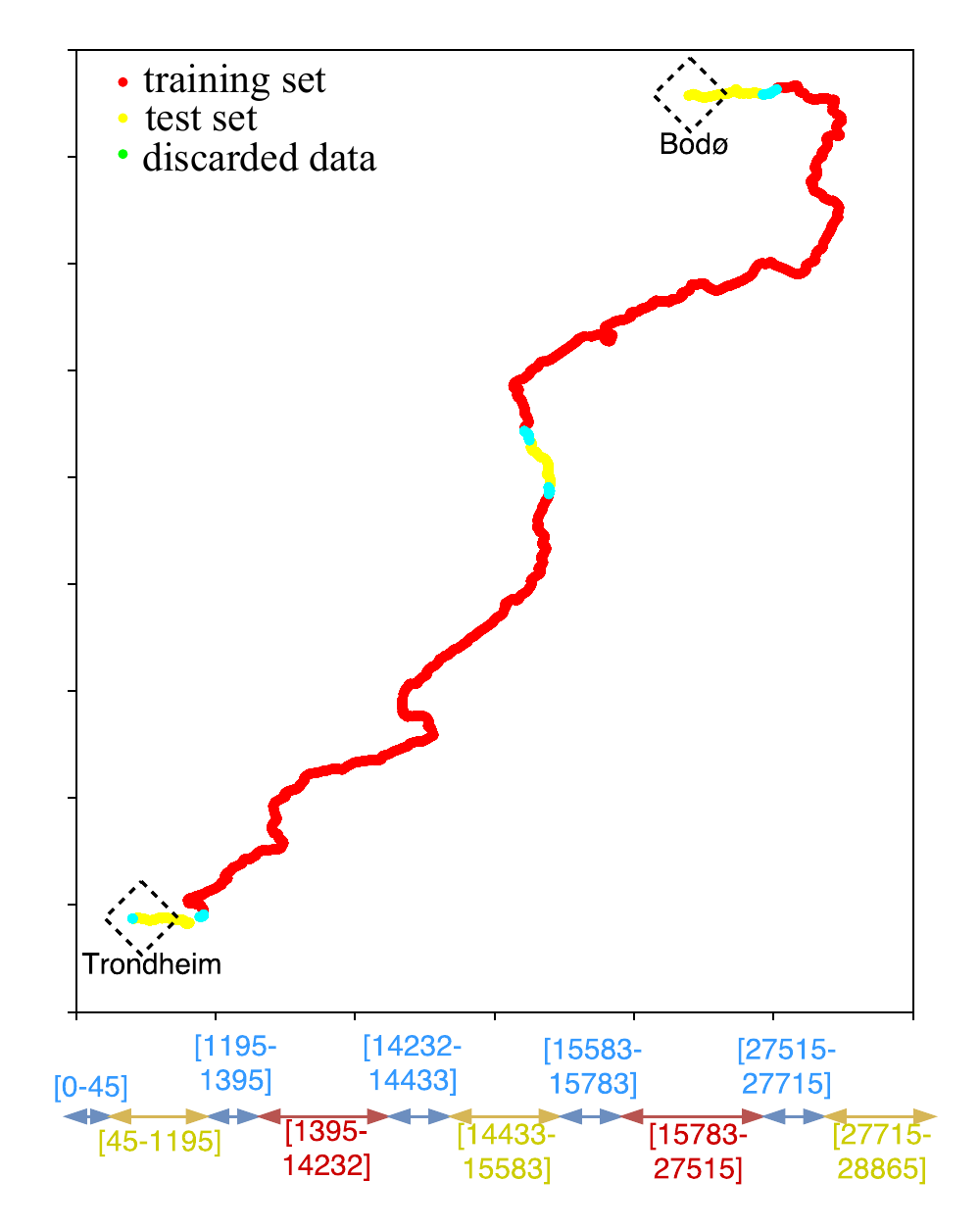}
	\caption{\label{fig:tramosnordland} 
    Proposed dataset partition for the Nordland dataset. \textbf{Top:} Geographical representation of the training (red) and test (yellow) sets. \textbf{Bottom:} Index representation of the distribution, w.r.t. frame index in the videos.
	}
\end{figure}

\section{The Nordland Dataset: Pre-Processing and Partitions}
\label{sec:dataset}

In this work, we have used the Nordland railroad videos.
In 2012, the Norway broadcasting company (NRK) made a documentary about the Nordland Railway, a railway line between the cities of Trondheim and Bodø. They filmed the $729$km journey with a camera in the front part of the train in winter, spring, fall and summer. The length of each video is about 10 hours and each frame is timestamped with the GPS coordinates. 

This dataset has been used by other research groups in place recognition, for example \cite{gomez2015training} and \cite{lowry2016supervised}. Each group uses different partitions for training and test, making difficult to reproduce the results. In this work we propose a specific partition of the dataset and a baseline, to guarantee a fair comparison between algorithms. Our intention is to release the processed dataset if the paper is accepted.

\subsection{Data pre-processing}
\label{sec:estructuradataset}
The first step, creating the dataset, was to extract the maximum number of images from each video. Moreover, GPS data corruption was fixed and we also eliminated tunnels and stations. After these steps, grabbing one frame per second, we obtained $28,865$ images per video. We used speed information from the GPS data to filter stations and a darkness threshold to filter tunnels.

\subsection{Dataset partitions}
\label{sec:estructuradatasetredes}

Fig. \ref{fig:tramosnordland} illustrates the partition of the whole image set in the Nordland dataset. We decided to create the test set with three different sequences of $1,150$ images (a total of $3,450$, yellow in the figure). The rest of the images were used for training ($24,569$, red in the figure). By using multiple sections, the variety of places and appearance changes contained in the test set increases. We also left a separation of a few kilometers between each test and train section by discarding some images in order to guarantee the difference between test and train data.

\subsection{Place labels}
\label{sec:lugares_datos_conjunto}
\begin{figure}[t!]
	\centering
	\includegraphics[width=.99\linewidth]{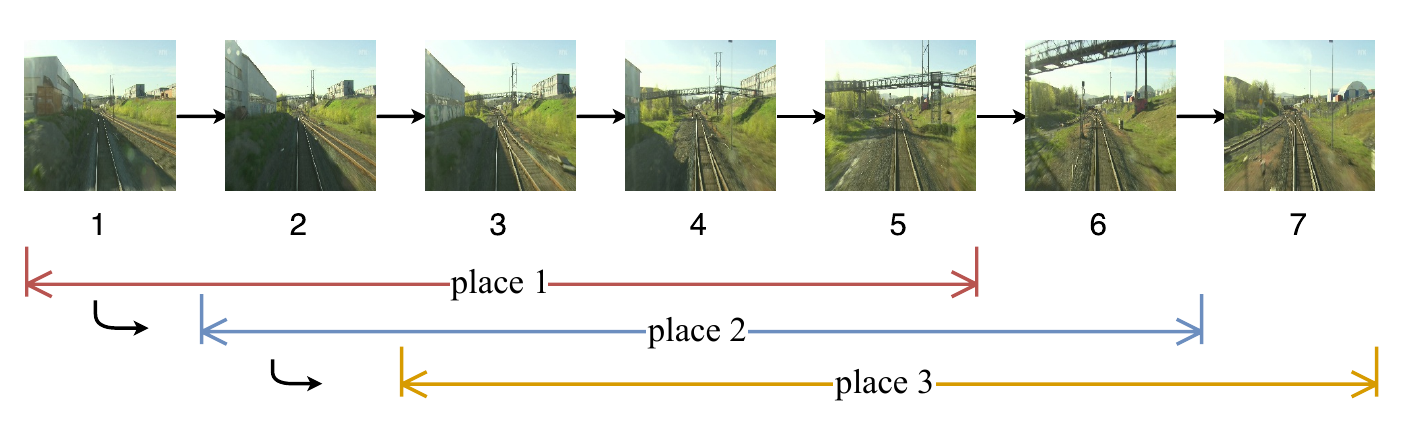}
	\caption{\label{fig:lugares_divididos}
    A sliding window of five images is considered in this work as the same place. Notice the similarity of consecutive images. The figure is best viewed in electronic format.
    } 
\end{figure}

Given the similarity between consecutive images, in this work we propose to consider that two images are of the same place if temporally they are separated by $3$ images or less. We applied a sliding window of $5$ images over the whole dataset in order to group images taken from five consecutive seconds. This process can be seen in Fig. \ref{fig:lugares_divididos}.

\section{Neural network architectures}
\label{sec:arquitecturas_neuronales}
\begin{figure}[t!]
	\centering
	\includegraphics[width=.99\linewidth]{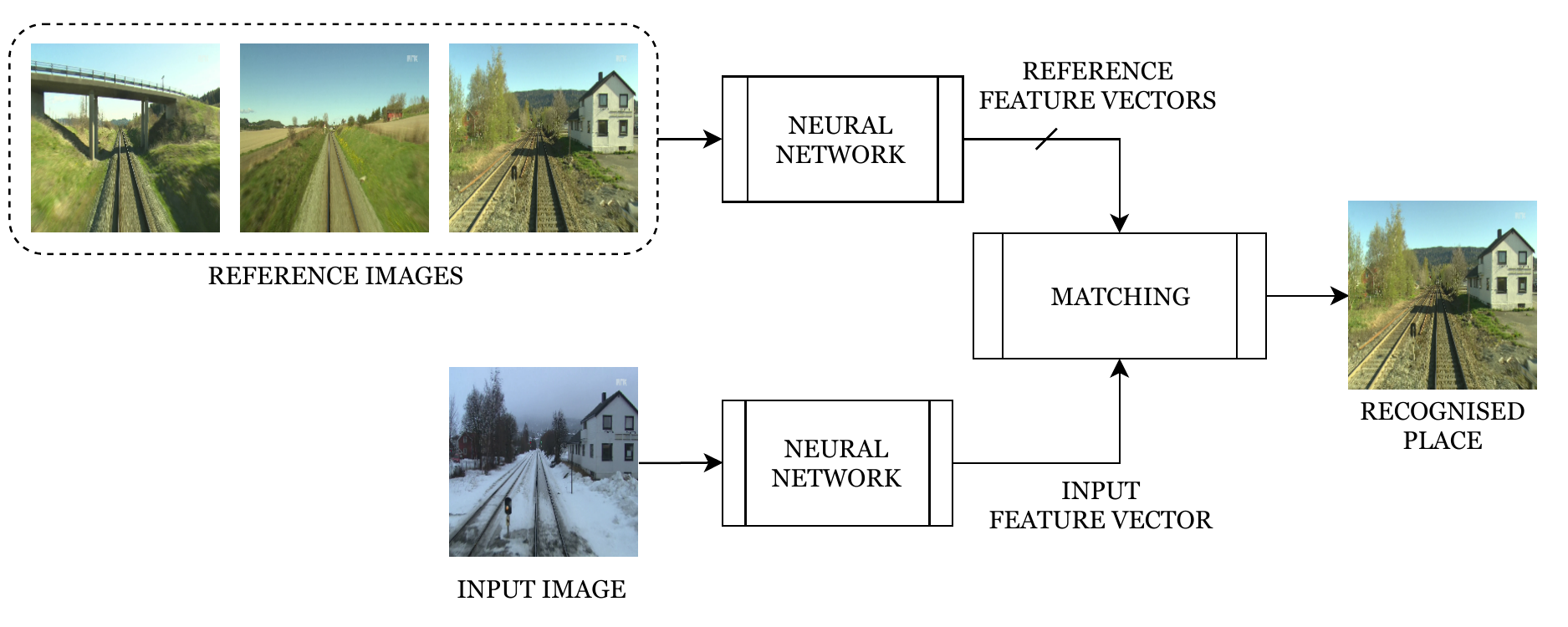}
    \caption[]{\label{fig:bloques_reconocedor} Overview of the place recognition algorithm. First, we extract a descriptor for every (visited place) image in the database. Second, for every new image (query) we extract its descriptor and compare it with those extracted from the database. The retrieved place will be the one with the most similar descriptor.
    }
\end{figure}

Fig. \ref{fig:bloques_reconocedor} shows the functional blocks of the proposed place recognition method. Our goal was to train a network to extract feature vectors that are close to the ones extracted from images of the same place, even in the presence of appearance changes. 
Our similarity metric is the Euclidean distance. We acknowledge that the distance function plays an important role in the feature space distribution, loss and optimization convergence. However, we preferred to focus our efforts on other parts of the problem rather than experimenting with other alternatives, e.g., the cosine distance. %

We studied three different ways of using neural networks. First of all, we evaluated the performance of features extracted by pre-trained networks. We then proceeded to train siamese and triplet architectures specifically for the problem of place recgonition.

\subsection{Pre-trained networks}
\begin{figure}[t]
	\centering
	\includegraphics[width=.99\linewidth]{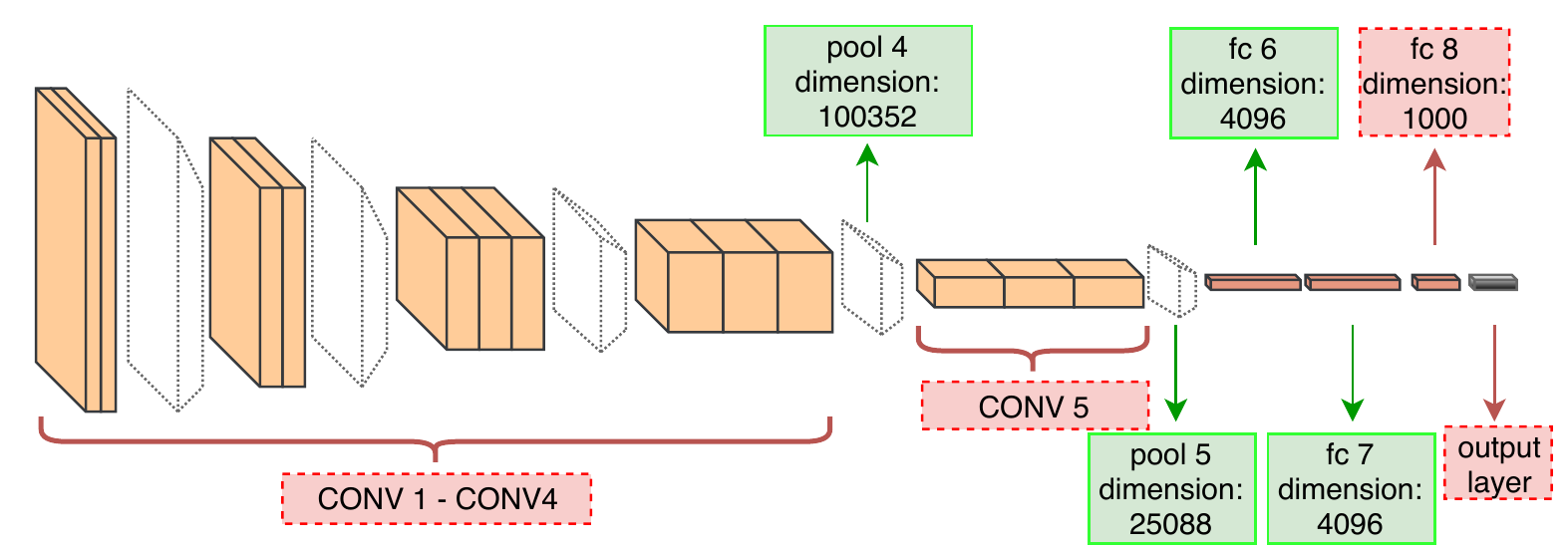}
	\caption{\label{fig:capas_vgg16_utilizadas}VGG-16 Layers. \textit{In red:} Layers not used. \textit{In green:} Used layers.}
\end{figure}

In \cite{DBLP:journals/corr/SunderhaufDSUM15}, S\"underhauf \textit{et al.} studied the performance of features from different neural networks for the purpose of place recognition. In this work, we analyzed the features extracted by some layers of the popular VGG-16 model \cite{simonyan2014very}, which was trained on Imagenet. Fig. \ref{fig:capas_vgg16_utilizadas} shows the structure of the model and the layers that we have evaluated. We have also evaluated the performance of the same architecture trained for scene recognition on the Places dataset. 

In the rest of this paper, by extracted feature vector we refer to the output of the neural network at the chosen layer after the non-linear activation. In the case of convolutional layers, we flattened the output tensor.
\begin{figure}[t!]
	\centering
	\includegraphics[width=.99\linewidth]{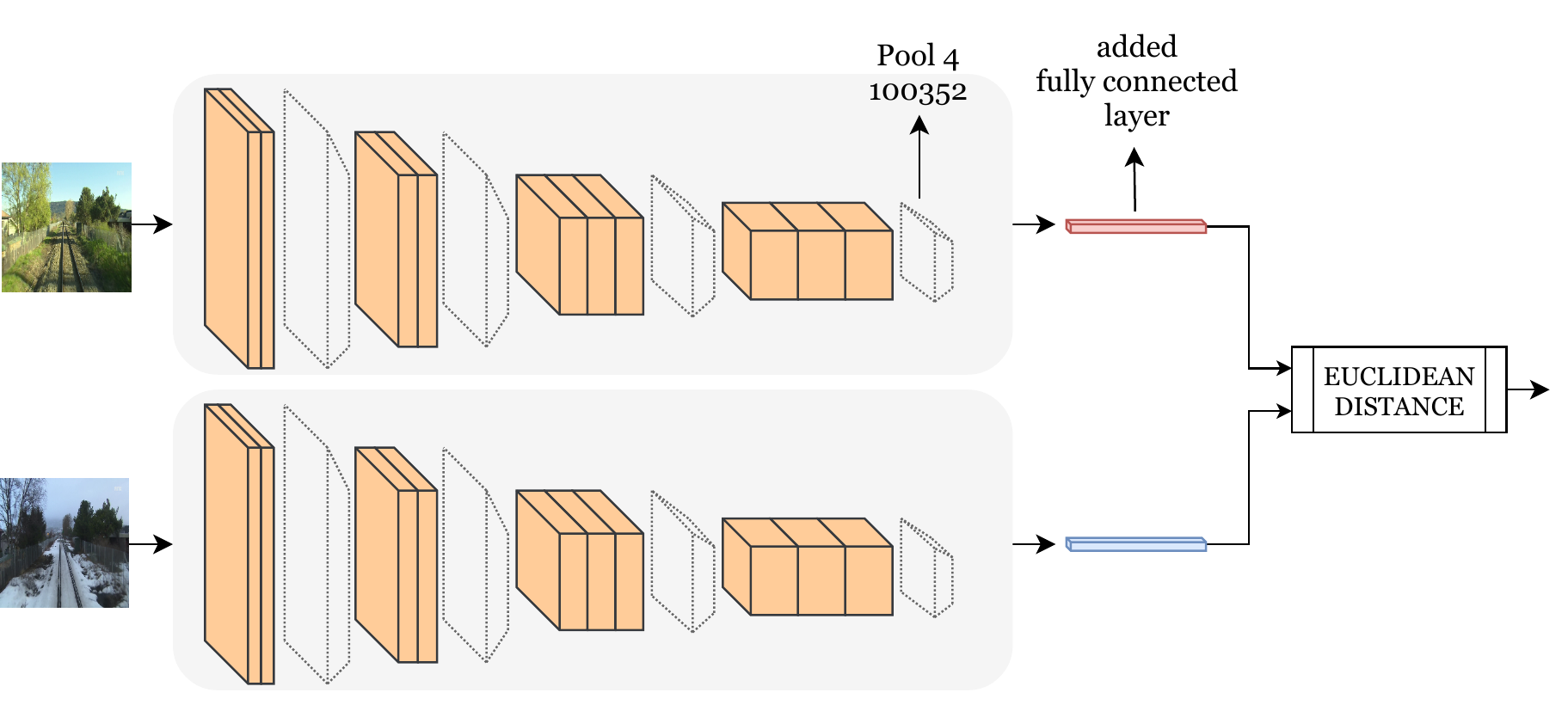}
	\caption{\label{fig:siamesa_lugar}Siamese architecture used in our work. We show in gray the pre-trained CNN blocks. The fully-connected layer added has $128$ neurons.} 
\end{figure}
\subsection{Siamese networks}
Siamese neural networks, proposed in \cite{bromley1994signature}, are capable of improving the robustness of pre-trained descriptors for place recognition. We modified the VGG-16 model in order to use a siamese architecture and added a new fully-connected layer (without activation function) after the one that showed the best performance in the pre-trained experiments. The final structure is showed in Fig. \ref{fig:siamesa_lugar}. Training was done for $5$ epochs with $834,746$ positive pairs (two images of the same place with different appearance) and $834,746$ negative pairs (two images of different places) taken from the previously mentioned training dataset. We used the contrastive loss \cite{hadsell2006dimensionality}. 
\subsection{Triplet networks}
As mentioned in Section \ref{sec:trabajo_relacionado}, G\'omez-Ojeda \textit{et al.} \cite{gomez2015training} were the first ones to train triplet networks with this purpose. Triplet neural networks improve the results of siamese architectures by training positive and negative pairs at the same time. Moving closer the descriptors from the same place and apart the descriptors from different places in the same instant leads to a more stable and efficient learning process. 

In order to use a triplet architecture, we modified the VGG-16 pre-trained model by adding a new fully-connected layer (without activation function) after the layer that performed better in the pre-trained experiments. We trained the new layer with $834,746$ image triplets for $5$ epochs. The loss function used in this case was the \emph{Wohlhart-Lepetit} loss. This loss, proposed in \cite{wohlhart2015learning} was also used in \cite{gomez2015training}:

\begin{equation}
E = max\left\{0, 1-\frac{d_n}{margin+d_p}\right\}
\end{equation}

Where $E$ is the loss error, $d_p$ is the distance between the positive and neutral input, $d_n$ is the distance between the neutral and negative input and $margin$ is a parameter that limits the difference between the distances.

In this function, the loss is zero when the positive pair is closer than the negative pair plus the margin. Moreover, the loss value is limited between 0 and 1. We set the margin value to 1 in all our experiments.

\section{Experimental Results}
\label{sec:resultados}
In order to evaluate our deep models, we used the images from one season as reference and images of a different season as query (summer against winter, winter against fall, etc.). Each image is processed by the neural network to produce the feature vector. After the extraction, each feature vector is compared with every feature vector of every reference season, and the closest one is considered the place predicted by the algorithm. This process is repeated for each one of the $3,450$ test images. The number of times that the closest place is the correct one gives the the fraction of correct matches fc, which is the metric that we have used. 

\begin{equation}
\text{fc} = \frac{\text{\# of correct predicted places}}{\text{\# of evaluated places}},
\end{equation}
It is important to note that we consider a match is correct when the closest feature vector corresponds to a place within a $5$-frames window. The distance between the feature vectors measures the confidence of the result and a distance threshold can be applied to obtain precision-recall curves. We have preferred to focus our analysis on the robustness of the extracted features. 

\subsection{Pre-trained}
\begin{figure}[t]
	\centering
	\includegraphics[width=.94\linewidth]{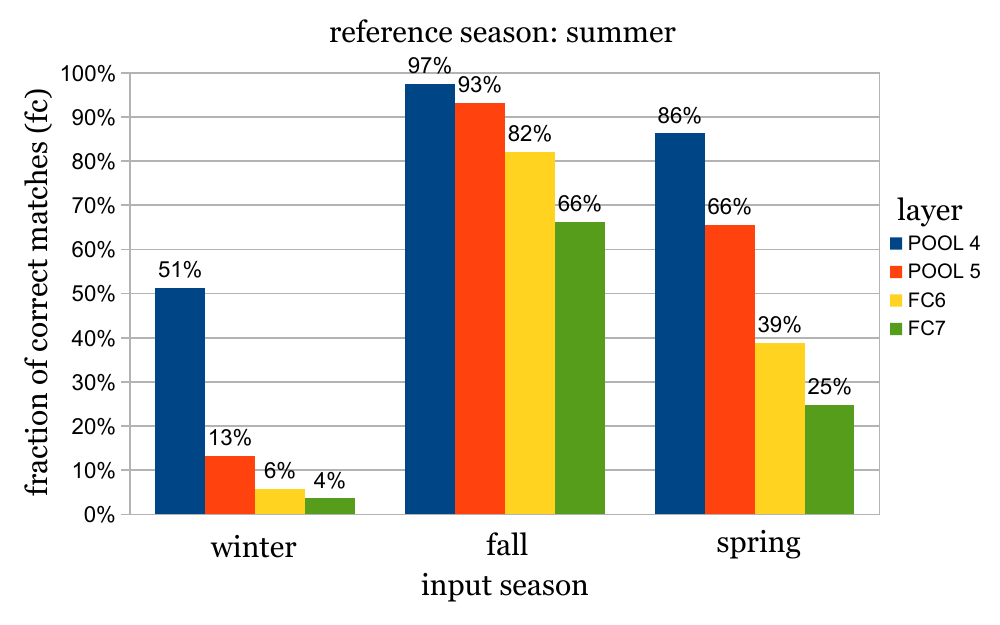}
	\caption{\label{fig:imagenet_capas}Fraction of correct matches using the pre-trained VGG-16 layers as feature extractors with summer as reference season and the other seasons as input.}
\end{figure}

Fig. \ref{fig:imagenet_capas} shows the results obtained from the original VGG-16 pre-trained model. Out of all the studied layers, we found that features extracted from the fourth pooling layer (\textit{pool}4) had the highest fraction of correct matches in all the season combinations. The results are worse as the layers are closer to the VGG-16 output. The main reason might be that, as the dimension of the layer decreases, some of the information that is robust to appearance changes is lost. Moreover, the last layers of the model contain semantic information which is specific to the original problem.

\begin{figure}[t]
	\centering
	\includegraphics[width=.94\linewidth]{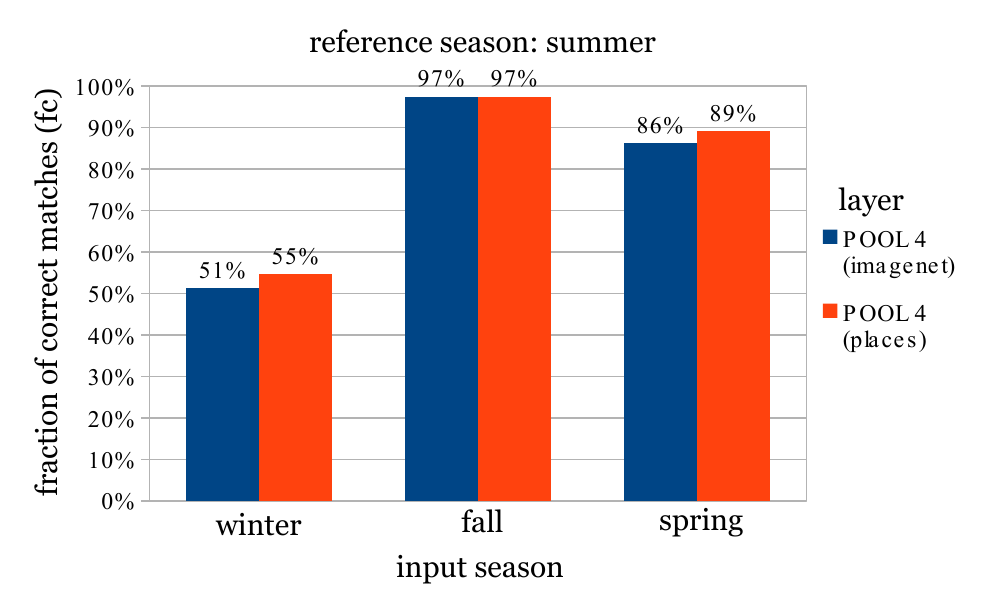}
	\caption{\label{fig:places_vs_imagenet}Fraction of correct matches using features from the \emph{pool4} layer of VGG-16. We compare the Imagenet pre-trained version vs the Places dataset pre-trained one. We show the results with summer as reference season and the other seasons as input.}
\end{figure}
\begin{figure}[t]
	\centering
	\includegraphics[width=.94\linewidth]{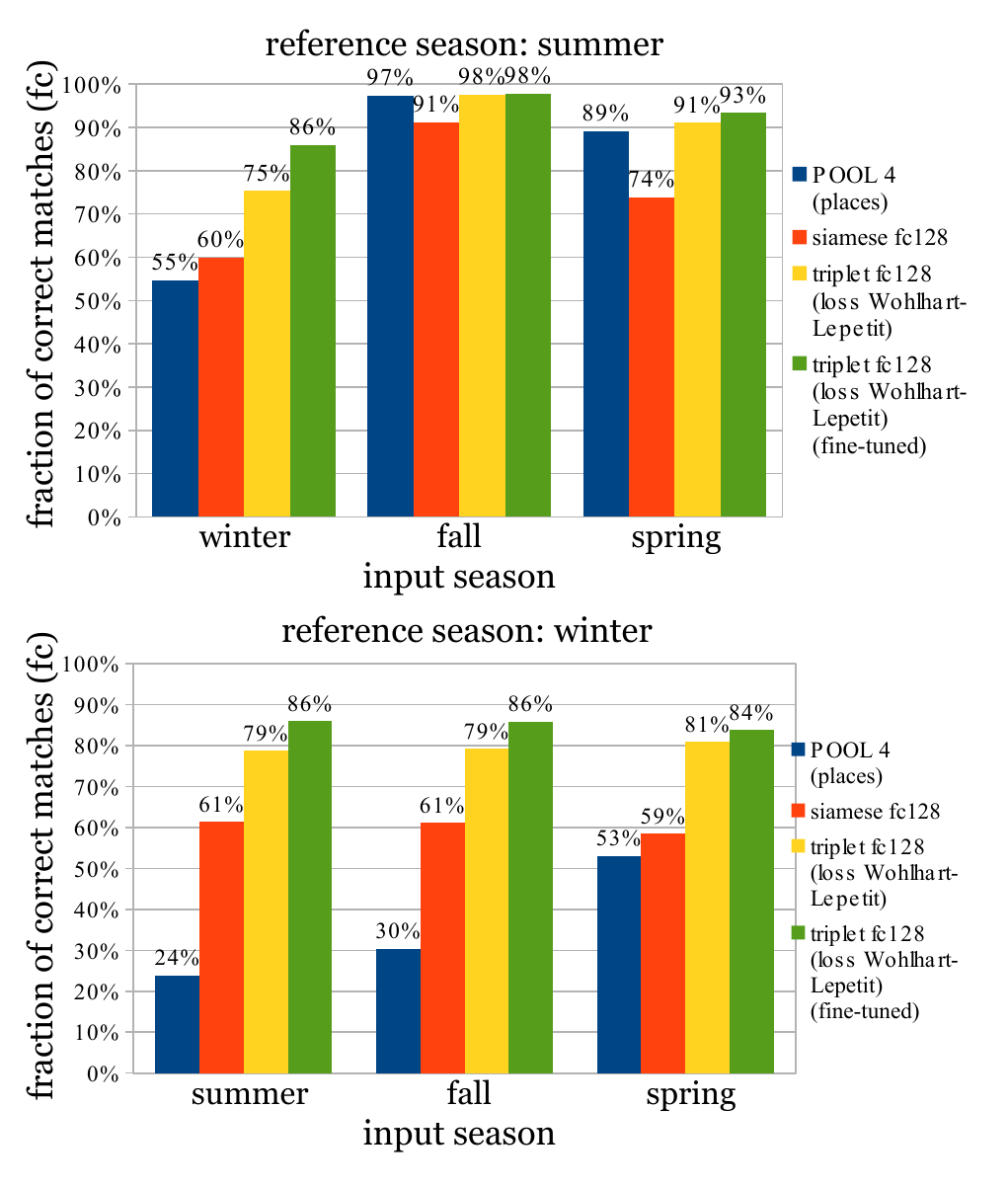}
	\caption{\label{fig:triplet_vs_siamesa_vs_places}Fraction of correct matches using different strategies: Pre-trained, siamese and triplet networks (with and without fine-tuning). \textit{Top:} Results with summer as reference season and the other seasons as input. \textit{Bottom:} Results with winter as reference season and the other seasons as input.}
\end{figure}
After that, we compared the performance of the original VGG-16 model to the VGG-16 model trained on the Places dataset by evaluating the features extracted from the \textit{pool}4 layer. We observed that the model trained for scene-recognition achieved better results in all the studied combinations, as shown in Fig. \ref{fig:places_vs_imagenet}. The main reason behind this is that, in order to classify scenes, the internal layers of the model have learned to extract features that are more useful for place recognition.  

In the rest of our experiments, we decided to use the fourth pooling layer of the VGG-16 trained on the Places dataset as the starting point. The extracted feature vectors have a dimension of $100,352$.

\subsection{Siamese and triplets}

After several experiments, we observed that a descriptor size of $128$ is sufficiently discriminative for place recognition. Increasing the size of the layer increases the computational cost without a significant improvement in the accuracy.

Fig. \ref{fig:triplet_vs_siamesa_vs_places} compares the results obtained with the pre-trained, siamese and triplet architectures. The pre-trained network only outperformed the siamese architecture in some combinations where summer images were used as reference. It should be noted that the siamese feature vector has $128$ dimensions, while the pre-trained one has $100,352$. Even if the siamese network has not outperformed all the pre-trained results, the siamese architecture has learnt to extract a much smaller feature vector, while keeping the discriminative information.

On the other hand, the triplet network outperformed the siamese and pre-trained models in all the studied combinations. The triplet results that we show in Fig. \ref{fig:triplet_vs_siamesa_vs_places}, belong to two different experiments. In our first experiments, we trained the newly added layer (\textit{triplet fc128 - loss Wohlhart Lepetit}). We then proceeded to train the layer while fine-tuning the rest of the VGG-16 pre-trained structure (\textit{triplet fc128 - loss Wohlhart Lepetit - fine-tuned}). It can be observed that the accuracy of the fine-tuned model is higher.

We conclude that the best results were obtained with the fine-tuned triplet network, starting from the weights of the pre-trained VGG-16-Places and adding a fully-connected layer with an output dimension of $128$.

Table \ref{table:fc_estaciones} shows the fraction of correct matches achieved for every possible combination of reference-input seasons. 
\begin{table}[t]
	\centering
	\caption{Fraction of correct matches for every season combination.}
	\label{table:fc_estaciones}
	\begin{tabular}{@{}cclll@{}}
		\toprule
		input \textbackslash reference & summer                   & fall                    & winter                 & spring                \\ \midrule
		summer                            & --- & 0.8548                   & 0.8591                   & 0.9545                  \\
		fall                             & 0.9777                   & --- & 0.8583                   & 0.9562                   \\
		winter                          & 0.8597                   & 0.9771                   & --- & 0.9545                   \\
		spring                         & 0.9336                   & 0.94                     & 0.8388                   & --- \\ \bottomrule
	\end{tabular}
\end{table}

\subsection{Comparison against other approaches}
\begin{figure}[t]
	\centering
	\includegraphics[width=.95\linewidth]{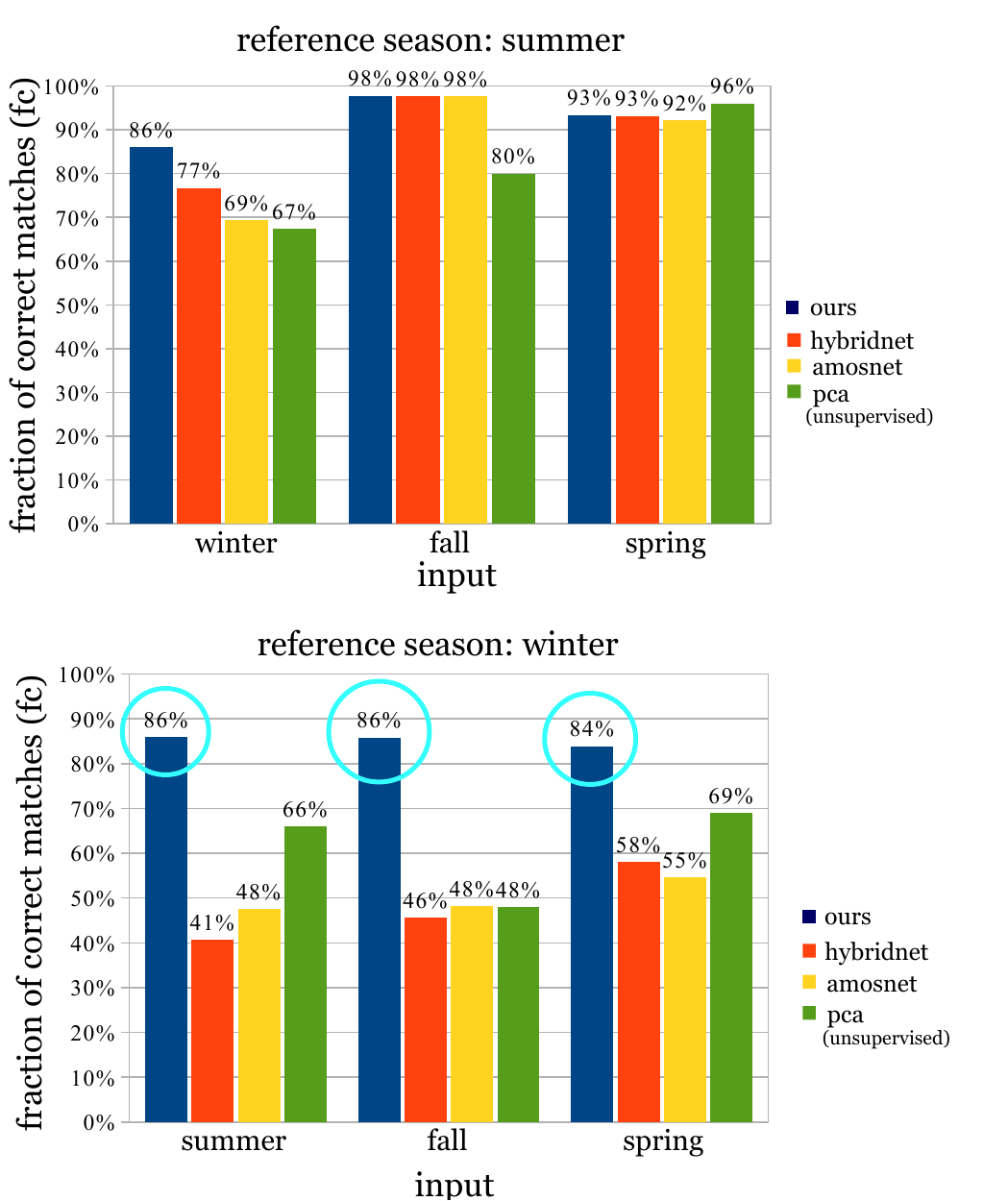}
	\caption{\label{fig:triplet_vs_amos_vs_hybrid_fc}Fraction of correct matches comparison: our work, Hybridnet, Amosnet and the unsupervised PCA technique. \textit{Top:} Results with summer as reference season and the other seasons as input. \textit{Bottom:} Results with winter as reference season and the other seasons as input.}
\end{figure}
Fig. \ref{fig:triplet_vs_amos_vs_hybrid_fc} shows the comparison between our results, the PCA technique of \cite{lowry2016supervised} and the two neural network models trained in \cite{chen2017deep} (Hybridnet and Amosnet). The comparison is made using summer and winter as the reference seasons. Notice that our model matches or outperforms the other techniques in almost every combination, and particularly in those with drastic appearance changes. In the most challenging cases (the ones with winter as reference) the best result is obtained with summer and fall as input seasons, where our model achieved 86\% of correct matches while the second best, the unsupervised PCA \cite{lowry2016supervised}, obtained less than 66\%. 

The unsupervised PCA results were obtained from their original paper \cite{lowry2016supervised}. For Hybridnet and Amosnet, we downloaded the models from the authors of \cite{chen2017deep} and tested their performance in the test partition of our dataset.

\begin{figure}[t!]
	\centering
	\includegraphics[width=.87\linewidth]{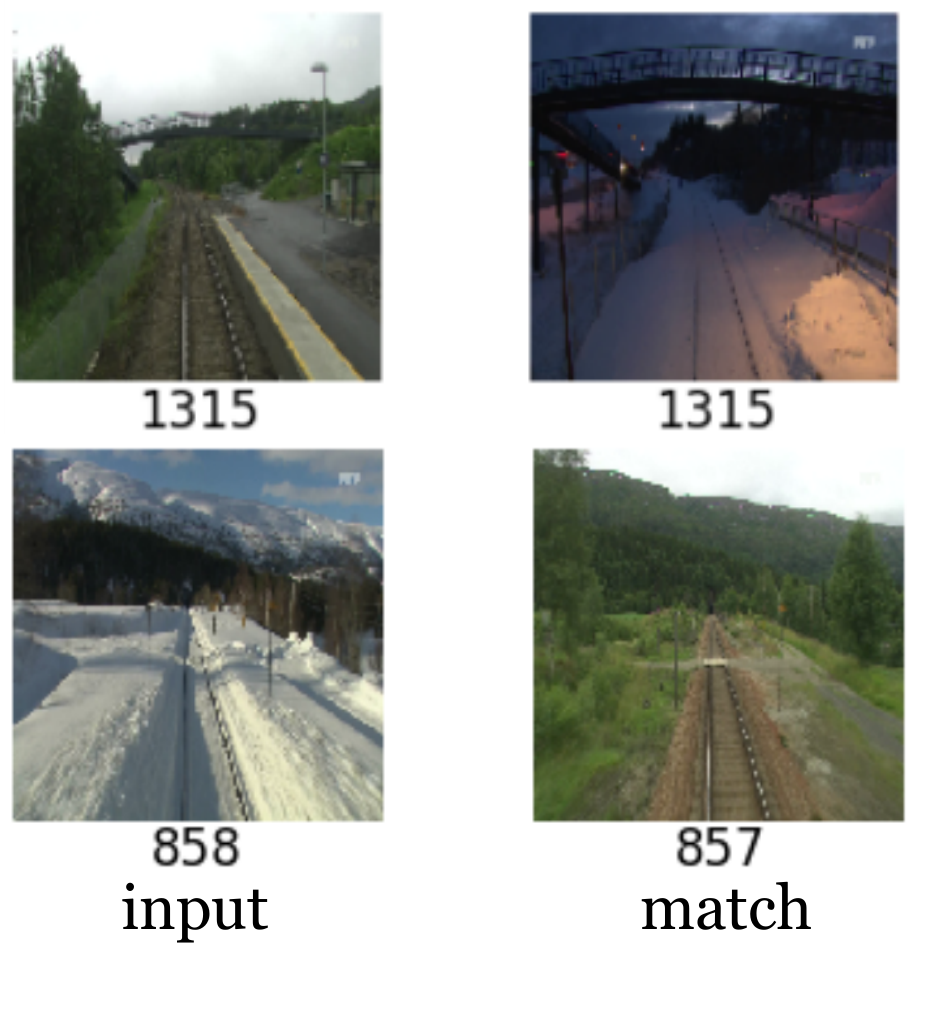}
	\caption{\label{fig:parejas_buenas_malas}Places correctly recognized by our algorithm. The index in the sequence is shown at the bottom of each image.}

\end{figure}

\begin{figure}[t!]
	\centering
	\includegraphics[width=.87\linewidth]{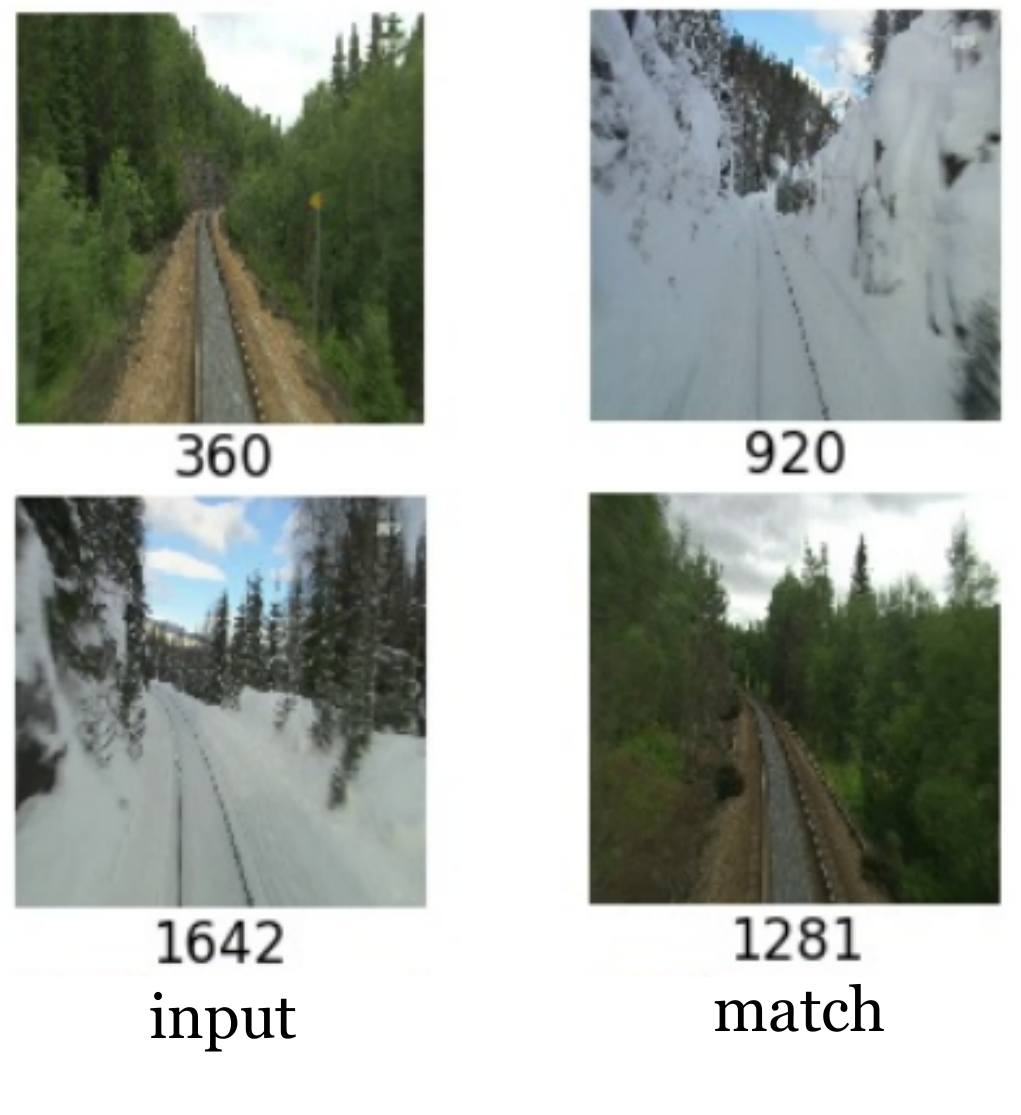}
	\caption{\label{fig:parejas_malas}False positive examples from our algorithm. The index in the sequence is shown at the bottom of each image. Notice that these particular places are difficult even for a human.}
\end{figure}

Finally, Fig. \ref{fig:parejas_buenas_malas} shows two examples of correct matches. Notice that our method is robust to strong changes produced by snow and illumination. Fig. \ref{fig:parejas_malas} shows two examples of incorrect matches. Notice that both are difficult even for a human. The similarities in the geographical features of those places make them look like the same place. 

\section{Conclusions}
\label{sec:conclusiones}
In this work we have implemented a place recognition method which is robust to appearance changes, in particular to those caused by weather conditions. Our proposal works by training a neural network to extract a descriptor, that can be compared with others using the Euclidean distance. 

Our experiments show that siamese and triplet neural networks learn robust features to appearance changes. Triplet neural networks achieved better results than siamese ones. We show that a VGG-16 model trained on the Places dataset shows a reasonable performance, improved by fine-tuning. Finally, we have shown that our method achieves state-of-the-art results in place recognition on the Nordland dataset.

\bibliographystyle{ieeetr}
\bibliography{tfgbib}

\end{document}